\PassOptionsToPackage{svgnames}{xcolor}
\documentclass{ecai}

\usepackage{latexsym}
\usepackage{amssymb}
\usepackage{amsmath}
\usepackage{amsthm}
\usepackage{booktabs}
\usepackage{enumitem}
\usepackage{graphicx}
\usepackage{color}

\usepackage{endnotes}
\usepackage{natbib}
\usepackage[noend]{algpseudocode}
\usepackage{algorithm}
\usepackage{mathtools}
\usepackage{tikz}
\usetikzlibrary{positioning, shapes.geometric, arrows.meta, calc}
\usepackage{subcaption}
\usetikzlibrary{decorations.markings}
\DeclareMathOperator*{\argmax}{arg\,max}
\DeclareMathOperator*{\argmaxhat}{arg\,\widehat{max}}
\DeclareMathOperator*{\eqdef}{\stackrel{\text{\tiny def}}{=}}
\usepackage[svgnames]{xcolor}
\usepackage{pgfplots}
\pgfplotsset{compat=1.17} 
\usepackage{filecontents} 
\usepackage{pgfplotstable} 
\usepackage{amssymb}
\usepackage{pifont}
\newcommand{\tick}{\checkmark}
\newcommand{\cross}{\ding{55}}

\newtheorem{theorem}{Theorem}
\newtheorem{lemma}[theorem]{Lemma}

\newtheorem{definition}{Definition}
\newtheorem{example}{Example}

\newcommand{\BibTeX}{B\kern-.05em{\sc i\kern-.025em b}\kern-.08em\TeX}

\begin{document}


\begin{frontmatter}

\paperid{9166} 

\title{Efficient Computation of Blackwell Optimal Policies using Rational Functions}

\author[A]{\fnms{Dibyangshu}~\snm{Mukherjee}}
\author[A]{\fnms{Shivaram}~\snm{Kalyanakrishnan}}

\address[A]{IIT Bombay, Mumbai, India \\
\{dbnshu, shivaram\}@cse.iitb.ac.in}


\begin{abstract}
Markov Decision Problems (MDPs) provide a foundational framework for modelling sequential decision-making across diverse domains, guided by optimality criteria such as discounted and average rewards. However, these criteria have inherent limitations: discounted optimality may overly prioritise short-term rewards, while average optimality relies on strong structural assumptions. Blackwell optimality addresses these challenges, offering a robust and comprehensive criterion that ensures optimality under both discounted and average reward frameworks. Despite its theoretical appeal, existing algorithms for computing Blackwell Optimal (BO) policies are computationally expensive or hard to implement.

In this paper we describe procedures for computing BO policies using an ordering of rational functions in the vicinity of $1$. We adapt state-of-the-art algorithms for deterministic and general MDPs, replacing numerical evaluations with symbolic operations on rational functions to derive bounds independent of bit complexity. For deterministic MDPs, we give the first strongly polynomial-time algorithms for computing BO policies, and for general MDPs we obtain the first subexponential-time algorithm. We further generalise several policy iteration algorithms, extending the best known upper bounds from the discounted to the Blackwell criterion.
\end{abstract}

\end{frontmatter}


\section{Introduction}

Markov Decision Problems (MDPs) are a widely used mathematical framework for modelling sequential decision-making problems. They form the backbone of reinforcement learning where agents learn to take decisions by interacting with an environment. Applications span diverse fields, including treatment planning in healthcare~\citep{RL-medicine}, automated control systems~\citep{RL-selfdriving}, robotics~\citep{RL-robotics}, game-solving~\citep{silver2016}, and financial portfolio management~\citep{RL-finance}. Their flexibility and rigorous foundation make MDPs a central tool in operations research, artificial intelligence, and economics.

An MDP is characterised by a tuple $(S,A,T,R)$, where $S$ is the set of states and $A$ is the set of actions. In this paper we assume $S$ and $A$ are finite with sizes $n$ and $k$ respectively. When an agent takes an action $a\in A$ from state $s \in S$, it transitions to a new state $s'$ with a probability $T(s,a,s')$ and receives a mean reward $R(s,a)$. The key feature of MDPs is the Markov Property, which asserts that the transition dynamics depend only on the current state and action and not on the sequence of past states. 

The objective in an MDP is to determine a policy: a rule that specifies the action to take in each state. Starting from an initial state $s_0$, if the agent follows a policy $\pi:S \to A$, it encounters a sequence $\left(s^t, \pi(s^t), r^t\right)_{t=0}^\infty$. The long-term reward of the agent for a state $s$ is determined by the chosen \textit{optimality criterion}.

Under the \textbf{discounted reward} criterion, a policy is evaluated using a value function that quantifies the cumulative discounted sum of rewards obtained by following the policy from state \(s\), expressed as:
\begin{equation} \label{discounted-expected-value}
    V_\gamma^\pi(s) = \lim_{T \to \infty} \mathbb{E}_\pi\left[ \sum_{t=0}^{T-1} \gamma^{t} R(s^t, a^t) \bigg| \ s^0 = s \right], 
\end{equation}  
where the discount factor $\gamma \in [0,1)$ is specified as part of the MDP.

The discounted framework is widely favoured~\cite{Sutton+Barto:1998,LeahyKSS22} for its mathematical simplicity, notably due to the contraction property. However, it is also subject to limitations, which we discuss later.

Under the \textbf{average reward} criterion, the value of a policy \(\pi\) for a state \(s\) is characterised by two components: the \textit{gain} and the \textit{bias}.  
The gain, denoted by \(V^{\pi}_{\text{g}}(s)\), represents the long-term average reward per time step under the policy \(\pi\) from state $s$. It is defined as:  
\[
V^{\pi}_{\text{g}}(s) = \lim_{T \to \infty} \mathbb{E}_{\pi}\left[ \frac{1}{T} \sum_{t=0}^{T - 1} R(s^{t}, a^{t}) \;\middle|\; s^{0} = s \right].
\] 
The bias, denoted by \(V^{\pi}_{\text{b}}(s)\), reflects the transient behaviour of the system, capturing how the reward dynamics evolve before the steady-state is reached. It is defined for each state $s$ as:  
\[
V^{\pi}_{\text{b}}(s) = \lim_{T \to \infty} \mathbb{E}_{\pi}\left[ \sum_{t=0}^{T - 1} \left( R(s^t, a^t) - V^{\pi}_{\text{g}}(s) \right) \;\middle|\; s^0 = s \right].
\]  

Gain and bias optimality are specific instances of a broader optimality criterion known as sensitive discount optimality~\citep{veinott1969}, which focuses on the cumulative sum of rewards as the discount factor approaches 1. Each of these criteria defines an optimal policy, $\pi^\star$, such that $V^{\pi^\star}$ \textit{dominates} (as a vector) $ V^\pi$ for all policies $\pi$. Here, $V^\pi$ serves as a placeholder for the value corresponding to the given optimality criterion, such as $V^{\pi}_{\text{g}}$ or $V^{\pi}_\gamma$.

MDPs can be solved, or an optimal policy found, using value iteration, policy iteration, or linear programming under both optimality criteria. These algorithms are straightforward and effective in the discounted case. However, in the average-reward case, the structure of the Markov chains induced by stationary policies plays a significant role, making it harder to design a general, simple algorithm. 

\citet{blackwell1962} introduced a refined notion of optimality, known as \textit{Blackwell optimality}. A policy $\pi$ is Blackwell Optimal (BO) if there exists a threshold $\overline{\gamma} \in (0,1)$ such that $\pi$ remains optimal under the discounted reward criterion for all discount factors $\gamma \in (\overline{\gamma}, 1)$. 
By definition, any BO policy is discount-optimal for all sufficiently large discount factors, and is also average-optimal.
Thus Blackwell optimality serves as a conceptual bridge between average and discounted optimality. While a BO policy is guaranteed to exist for every finite MDP~\citep{blackwell1962}, existing methods for computing BO policies are either inefficient or overly intricate (see sections \ref{section:BO} and \ref{section:Lit Review}). This paper proposes simple and efficient techniques for computing BO policies.

We develop a symbolic method for ordering rational functions near the point \(x = 1\). This idea originates from the work of~\citet{hordijk-kallenberg}, who applied such orderings within the linear programming framework as part of a simplex-based method for solving MDPs over an entire range of discount factors. In contrast, we incorporate symbolic ordering directly into the dynamic programming framework. By treating the discount factor as a symbolic variable, we express value and action-value functions as rational functions and use their relative orderings to guide policy improvement. This enables us to reinterpret and extend classical algorithms---such as policy iteration---for computing BO policies. Crucially, our approach yields algorithms with provable efficiency and establishes the tightest known bounds to date for computing BO policies, independent of the input's bit representation.

\subsection{Contributions}
We use our ordering of rational functions to simulate the trajectory of various algorithms on an MDP with a sufficiently large discount factor. This methodology enables the following contributions:
\begin{itemize}

    \item 

    \citet{post2013simplex} showed that the Max-Gain simplex algorithm converges to the optimal policy in strongly polynomial time for deterministic MDPs, with a bound of \(O(n^5k^2 \log^2 n)\) iterations.

    \smallskip

    \citet{madani2010discounted} extended the classical algorithm of \citet{karp1978} to the discounted setting and achieved a bound of \(O(n^2k)\) for solving deterministic MDPs.

    \smallskip

    We generalise both of these algorithms to the Blackwell setting, preserving their respective bounds up to a polynomial factor. This yields the first strongly polynomial guarantees for computing Blackwell-optimal policies in deterministic MDPs.

    \medskip
    
    \item We obtain the first direct policy improvement procedure for computing BO policies that does not rely on Laurent series expansions. When combined with the Random-Facet algorithm~\citep{kalai,MSW}, this approach yields a subexponential expected bound of \(\text{poly}(n,k) \cdot \exp(O(\sqrt{n\log n}))\) for general MDPs---the tightest known bound to date that is independent of the bit-size of the input.

    \medskip

    \item The switching rule used in policy iteration plays a crucial role in determining the algorithm’s complexity. We analyse three switching rules that achieve the tightest known upper bounds for discounted MDPs, and generalise each to the Blackwell setting while preserving their bounds up to polynomial factors. 

    \medskip

     \item For every MDP, there exists a threshold discount factor beyond which all discount-optimal policies are also Blackwell-optimal. A tight upper bound on this threshold facilitates the computation of BO policies, while a large lower bound highlights the inherent complexity of the problem and the limitations of certain algorithmic approaches discussed in section~\ref{section:Summary}.

     \smallskip 
     We construct an MDP whose threshold discount factor is exponentially close to 1, thereby establishing the best-known lower bound on this threshold.

\end{itemize}

The next section introduces Blackwell optimality, providing background, motivation, and outlining the key challenges in computing BO policies. Section~\ref{section:Lit Review} reviews the relevant literature. Section~\ref{section: Rational Fn} presents our rational function ordering framework, which forms the basis for the algorithms developed in Section~\ref{section:Algos}. Finally, Section~\ref{section:Summary} concludes with a discussion and summary of our contributions.

\section{Blackwell Optimality} \label{section:BO}
The value function of a policy $\pi$, defined in~\eqref{discounted-expected-value}, satisfies the recursive \textit{Bellman equations}, which for $s \in S$ take the form:
\begin{equation} \label{bellman-V}
    V^\pi(s) = R(s, \pi(s)) + \gamma \sum_{s'} T(s, \pi(s), s') V^\pi(s').  
\end{equation}  
Similarly, for $s \in S, a \in A$, the action-value function is defined as:
\begin{equation} \label{bellman-Q}
    Q^\pi(s, a) = R(s,a) + \gamma \sum_{s'} T(s, a, s') V^\pi(s').  
\end{equation} 

\noindent
\textbf{Notation.} We write $V^\pi_\gamma$ and $Q^\pi_\gamma$ to make the dependence on $\gamma$ explicit, particularly in contexts where $\gamma$ may not be fixed.

For any policy \(\pi\), let \(P^\pi\) denote the \(n \times n\) stochastic matrix with entries \(P^\pi(s, s')=T(s, \pi(s), s')\). Similarly, let \(\mathbf{r}^\pi\) denote the $n \times 1$ reward vector with components \(R(s, \pi(s))\). The Bellman equation for the value vector can be expressed as:
\begin{equation} \label{bellman-V-vec}
    \mathbf{v}^\pi = \mathbf{r}^\pi + \gamma P^\pi \mathbf{v}^\pi.  
\end{equation}

Solving this system using Cramer's rule, the value of a state \(s\) under policy \(\pi\) is given by: 
\begin{equation} \label{value:cramer}
    \mathbf{v}^\pi_s = \frac{\mathbf{n}^\pi_s}{{d}_\pi},
\end{equation}
where: \({d}_\pi = |I - \gamma P^\pi|\) is the determinant of \(I - \gamma P^\pi\), and  \(\mathbf{n}^\pi_s\) is the determinant of the matrix formed by replacing the \(s\)-th column of \(I - \gamma P^\pi\) with \(\mathbf{r}^\pi\).  
Similarly, the action-value function can be written in vectorised form using the reward vector $\mathbf{r}_a$ and transition matrix $P_a$ for action $a$, with $\mathbf{v}^\pi$ substituted from Equation~\eqref{value:cramer}:
\begin{equation} \label{poly-Q}
    \mathbf{q}^\pi_a = \mathbf{r}_a + \gamma P_a \frac{\mathbf{n}^\pi}{{d}_\pi}.  
\end{equation}  

\subsection{BO Policies and Threshold Discount Factor}
\begin{definition}
    A policy $\pi$ is Blackwell-optimal if there exists $\gamma' \in [0,1)$ such that $V^\pi_\gamma(s) \geq V^{\pi'}_{\gamma}(s), \ \forall s \in S, \forall \pi' \in \Pi$ and $\gamma \in [\gamma',1)$.
\end{definition}
A BO policy is both discounted (as $\gamma \to 1$) and average-optimal, but the converse need not hold. See Figure~\ref{fig:mdp-BO} for an example. 
\begin{theorem}[\citet{blackwell1962}]
    Every finite MDP has at least one BO policy.
\end{theorem}
The proof of this result becomes evident through the developments presented in this paper. See section~\ref{subsection:application}. 

Threshold discount factors play an important role in computation of BO policies. We define two existing threshold discount factors below.
Let \(\Pi^\star_\text{bw}\) denote the set of BO policies and $\Pi^\star_{\gamma}$ the set of discount-optimal policies with discount factor $\gamma$. The Blackwell discount factor \(\gamma_{\text{bw}}\), introduced by~\citet{Clement-Petrik}, is defined as:  
\[
\gamma_{\text{bw}} \eqdef \inf\bigg\{\gamma \in [0,1) \ \bigg | \ \forall \gamma' \in (\gamma,1),\Pi^\star_{\gamma'}=\Pi^\star_{\text{bw}} \bigg\}.
\]
That is, $\gamma_{\text{bw}}$ is the smallest discount factor beyond which the set of discount-optimal policies coincides with the set of BO policies. It is guaranteed to exist in every finite MDP~\citep{Clement-Petrik}.

\citet{mukherjee-kalyanakrishnan} define a stronger threshold condition, requiring that the ordering of Q-values for every policy remains invariant beyond the threshold. Formally:
\begin{multline*}
    \gamma_{\text{Q}} \eqdef \sup\limits_{\substack{\pi \in \Pi; s \in S; a, a^{\prime} \in A}} \bigg\{ \inf \bigg\{ \gamma \in [0, 1) \ \bigg | \ \forall \tau \in (\gamma, 1), \\
    \left( Q^{\pi}_{\gamma}(s, a) > Q^{\pi}_{\gamma}(s, a^{\prime}) 
    \implies Q^{\pi}_{\tau}(s, a) > Q^{\pi}_{\tau}(s, a^{\prime}) \right) \bigg\} \bigg\}.
\end{multline*}
It can be shown that $\gamma_{\text{Q}}$ exists for every MDP, and that $\gamma_{\text{bw}} \leq \gamma_{\text{Q}}$~\cite{mukherjee-kalyanakrishnan}.

\subsection{The Case for BO Policies}
Most Reinforcement Learning problems aim to maximise the cumulative sum of rewards for an agent. In infinite-horizon tasks without absorbing goal states, this sum may diverge unless rewards are discounted. Discounted optimality, therefore, is widely adopted and applied in domains such as obstacle avoidance for robots~\citep{Mahadevan-Connell} and routing automated guided vehicles to serve multiple queues~\citep{Tadepalli-Ok}. Despite its appeal, discounted optimality can yield suboptimal behaviour, favouring short-term mediocre rewards over more valuable long-term outcomes. For such continuing tasks, average reward is often a more suitable objective. However, algorithms for average optimality and their convergence guarantees usually rely on strong structural assumptions about the underlying Markov chain, such as unichain or ergodicity, which can be restrictive and hard to verify~\citep{Tsitsiklis-2007}.

\begin{figure}[b]
\vspace{-0.1cm}
  \centering
  \begin{subfigure}[b]{0.23\textwidth}
    \centering
    \raisebox{5pt}{
    \begin{tikzpicture}[scale=0.75, 
      >={Stealth[length=5pt,width=6pt]},
      every edge/.append style={thick},
      node style/.style={draw, circle, minimum size=18pt, inner sep=0pt, align=center},
      red edge/.style={draw=DarkRed!60},
      blue edge/.style={draw=DarkBlue!70},
      grey edge/.style={draw=Black!65}
      ]
      \node[node style, fill=lightgray!20] (A) at (0,0) {$s_1$};
      \node[node style, fill=lightgray!20] (B) at (0,-1.4) {$s_2$};
      \node[node style, fill=lightgray!20] (C) at (0,-2.8) {$s_3$};

      \path[->, blue edge, dashed] (A) edge node[midway, xshift=-1.5, left, DarkBlue] {\scriptsize $5$ }(B);
      \path[->, grey edge] (B) edge node[midway, xshift =-1.5, left, Black!75] {\scriptsize $5$ }(C);
      \path[->, red edge, dotted, bend right=60] (A) edge node[left, DarkRed] {\scriptsize $10$ }(C); 
      \path[->, grey edge, bend left=40] (A) edge node[right, Black!75] {\scriptsize $0$ }(B); 
      \path[->, grey edge, loop, out=50, in=0, min distance=6mm] (C) edge node[right, Black!75] {\scriptsize $0$} (C);
    \end{tikzpicture}
    } 
    \caption{\hspace*{-10pt}}
    \label{fig:mdp-BO}
  \end{subfigure}
  \hfill
  \begin{subfigure}[b]{0.23\textwidth}
    \centering
    \begin{tikzpicture}[scale=0.25, every node/.style={scale=1},red edge/.style={draw=DarkRed!60},
    blue edge/.style={draw=DarkBlue!70}]
            \def\radius{4.5cm}
            \tikzset{>={Stealth[length=6pt,width=7pt]}}
        
            \node[draw, circle, fill=black, inner sep=2.5pt, label={[label distance=-0.4mm]90:$s_1$}] (A) at (90:\radius) {};
            \node[draw, circle, fill=black, inner sep=2.5pt, label={[label distance=-1.5mm]150:$s_2$}] (B) at (150:\radius) {};
            \node[draw, circle, fill=black, inner sep=2.5pt, label={[label distance=-1mm]210:$s_3$}] (C) at (210:\radius) {};
            \node[draw, circle, fill=black, inner sep=2.5pt, label={[label distance=-0.5mm]270:$s_4$}] (D) at (270:\radius) {};
            \node[draw, circle, fill=black, inner sep=2.5pt, label={[label distance=-1.5mm]330:$s_5$}] (E) at (330:\radius) {};
            \node[draw, circle, fill=black, inner sep=2.5pt, label={[label distance=-1.5mm]30:$s_6$}] (F) at (30:\radius) {};

            \draw[red edge, thick, postaction={decorate, decoration={markings, mark=at position 0.5 with {\arrow{>}}}}] 
                (A) arc (90:150:\radius) node[midway, DarkRed, above left, font=\scriptsize] {1};
            \draw[red edge, thick, postaction={decorate, decoration={markings, mark=at position 0.5 with {\arrow{>}}}}] 
                (B) arc (150:210:\radius) node[midway, DarkRed, left, font=\scriptsize] {1};
            \draw[red edge, thick, postaction={decorate, decoration={markings, mark=at position 0.5 with {\arrow{>}}}}] 
                (C) arc (210:270:\radius) node[midway, DarkRed, below left, font=\scriptsize] {0};
            \draw[red edge, thick, postaction={decorate, decoration={markings, mark=at position 0.5 with {\arrow{>}}}}] 
                (D) arc (270:330:\radius) node[midway, DarkRed, below right, font=\scriptsize] {3};
            \draw[red edge, thick, postaction={decorate, decoration={markings, mark=at position 0.5 with {\arrow{>}}}}] 
                (E) arc (330:390:\radius) node[midway, DarkRed, right, font=\scriptsize] {8};
            \draw[red edge, thick, postaction={decorate, decoration={markings, mark=at position 0.5 with {\arrow{>}}}}] 
                (F) arc (390:450:\radius) node[midway, DarkRed, above right, font=\scriptsize] {7};

            \draw[blue edge, dashed, thick, postaction={decorate, decoration={markings, mark=at position 0.5 with {\arrow{>}}}}] 
                (A) -- node[auto, DarkBlue, above, fill=white, xshift=-4pt, yshift=-2pt, inner sep=0.5pt, outer sep=0pt, font=\scriptsize] {6} (C);
            \draw[blue edge, dashed,thick, postaction={decorate, decoration={markings, mark=at position 0.5 with {\arrow{>}}}}] 
                (B) -- node[auto, DarkBlue, above, fill=white, xshift=-3pt, yshift=-6pt, inner sep=0.5pt, outer sep=0pt, font=\scriptsize] {8} (D);
            \draw[blue edge, dashed, dashed, thick, postaction={decorate, decoration={markings, mark=at position 0.5 with {\arrow{>}}}}] 
                (C) -- node[auto, DarkBlue, above, fill=white, xshift=1pt, yshift=-7pt, inner sep=0.5pt, outer sep=0pt, font=\scriptsize] {3} (E);
            \draw[blue edge, dashed,thick, postaction={decorate, decoration={markings, mark=at position 0.5 with {\arrow{>}}}}] 
                (D) -- node[auto, DarkBlue, above, fill=white, xshift=4pt, yshift=-4pt, inner sep=0.5pt, outer sep=0pt, font=\scriptsize] {9} (F);
            \draw[blue edge, dashed, thick, postaction={decorate, decoration={markings, mark=at position 0.5 with {\arrow{>}}}}] 
                (E) -- node[auto, DarkBlue, above, fill=white, xshift=3pt, yshift=1pt, inner sep=0.5pt, outer sep=0pt, font=\scriptsize] {8} (A);
            \draw[blue edge, dashed, thick, postaction={decorate, decoration={markings, mark=at position 0.5 with {\arrow{>}}}}] 
                (F) -- node[auto, DarkBlue, above, fill=white, xshift=-1pt, yshift=2pt, inner sep=0.5pt, outer sep=0pt, font=\scriptsize] {4} (B);
    
        \end{tikzpicture}
    \caption{\hspace*{-5pt}}
    \label{fig:mdp-fib}
  \end{subfigure}
  \vspace{0.3cm}
  \caption{\footnotesize \textbf{a)} Example of a DMDP with 3 states and 3 actions: action 1 (solid), action 2 (dashed), and action 3 (dotted). From states $s_2$ and $s_3$, all actions produce the same effect. Consider policies $\pi_1 = (1, 1, 1)$, $\pi_2 = (2, 1, 1)$, and $\pi_3 = (3, 1, 1)$. All three policies yield the same gain, $V^{\pi_i}_{\text{g}}(s_1) = 0$. Yet, their biases differ: $V^{\pi_2}_{\text{b}}(s_1) = V^{\pi_3}_{\text{b}}(s_1) = 10$, while $V^{\pi_1}_{\text{b}}(s_1) = 5$. Their discounted values are:
    $V^{\pi_1}_{\gamma}(s_1) = 5\gamma, V^{\pi_2}_{\gamma}(s_1) = 5 + 5\gamma, V^{\pi_3}_{\gamma}(s_1) = 10$.
    Since $0 < \gamma < 1$, it follows that $V^{\pi_1}_{\gamma}(s_1) < V^{\pi_2}_{\gamma}(s_1) < V^{\pi_3}_{\gamma}(s_1)$. Note that only $\pi_{3}$ is BO. \textbf{b)} Example of an DMDP with $S=\{s_i\}_{i=1}^6$ and $A=\{a_1\text{ (solid)},a_2\text{ (dashed)}\}$. Transitions are deterministic; rewards are shown on edges. While the actual Blackwell discount factor is \(\gamma_{\text{bw}} = 0.8541\), the upper bound of ~\citet{Clement-Petrik} is: \(U=1 - \frac{1}{2.19 \times 10^{66}}\).}
  \label{fig:side-by-side}
\end{figure}
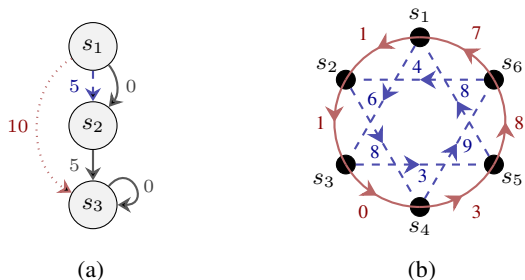

Consider the deterministic MDP in Fig.~\ref{fig:mdp-BO}, which admits three policies: $\pi_1$, $\pi_2$, and $\pi_3$. All three are gain-optimal, $\pi_2$ and $\pi_3$ are also bias-optimal, but only $\pi_3$ is Blackwell-optimal. This example shows that Blackwell optimality is strictly stronger than either gain or bias optimality---it selects policies that remain optimal for all sufficiently high discount factors, rather than only at a particular value or in the average-reward limit. Consequently, it provides a more robust and stable decision rule, especially when the discount factor is unknown, ill-defined, or subject to change. Moreover, in transient states, BO policies prioritise early reward collection, making them particularly effective when such states yield large immediate returns.

By unifying the strengths of the discounted and average‐reward criteria, Blackwell optimality offers a versatile framework for decision‐making. Computing a BO policy automatically yields an average‐optimal policy, which has a wide range of real‐world applications, including queuing networks~\citep{al2025simulation}, scheduling~\citep{Aydin-Oztemel}, inventory management~\citep{Das1999}, and transfer lines~\citep{MahadevanT98}. Accordingly,~\citet{Dewanto2020} identify advancing our understanding of Blackwell optimality as one of the pressing questions in reinforcement learning.

\subsection{Challenges in Computing BO Policies}
Algorithms for computing BO policies are either computationally expensive~\citep{veinott1969} or intricate and difficult to implement~\citep{sullivan-veinott}. 
To address this, \citet{Clement-Petrik} derive an upper bound $U$ on the threshold discount factor $\gamma_{\text{bw}}$, guaranteeing that for any $\gamma \geq U \geq \gamma_{\text{bw}}$, every discount-optimal policy is also BO. Their bound has the form
$U = 1 - \frac{1}{O(n^n) \cdot r_\infty \cdot m^n},$
where $m$ is the bit-size of the MDP instance and $r_\infty$ the maximum absolute reward.

 While this gives a direct recipe for computing BO policies for MDPs with rational data of finite precision, the bound is so conservative as to be impractical. For example, in Fig.~\ref{fig:mdp-fib}, the true threshold is $\gamma_{\text{bw}} \approx 0.85$, but the bound yields $U \approx 1 - O(10^{-66})$, a vast gap that limits its applicability.

In practice, the absence of tight bounds leads practitioners to use very large discount factors to approximate BO policies. To be safe, they may choose an extremely high $\gamma$ and solve the problem via policy iteration, value iteration, or linear programming. However, as $\gamma \to 1$, convergence slows dramatically~\citep{mahadevan1996b}, the Bellman matrix becomes nearly singular, and solving the resulting system of equations becomes numerically unstable. For large MDPs, the combination of high $\gamma$ and sparse transitions further exacerbates instability, increasing the risk of computational errors.

Our experiments show that value iteration slows markedly as the discount factor $\gamma$ approaches 1, with runtime scaling roughly as $\frac{1}{1 - \gamma}$. 
In contrast, both policy iteration and linear programming fail to converge beyond certain threshold values of $\gamma$.
We implemented VI and PI using the Python package \texttt{MDPtoolbox}~\cite{toolbox} with default parameters, and formulated the LP approach using the \texttt{cvxpy} library. 
Figure~\ref{fig:mdp-fail} presents results for a simple 2-state MDP with deterministic transitions, indicating the $\gamma$ ranges where each method either slows dramatically or fails. The table summarises these failures, showing that none of the algorithms can compute the discount-optimal policy for $\gamma > \gamma_{\text{fail}}$. Consequently, when $\gamma_{\text{bw}} > \gamma_{\text{fail}}$, these methods cannot produce the BO policy. In our implementations, $\gamma_{\text{fail}} \approx 1 - 10^{-17}$ for all three algorithms. Our code and data are available at~\cite{blackwell-pi-code}.

To illustrate the impact of this limitation, we examine two families of MDP instances with $\gamma_{\text{bw}} > \gamma_{\text{fail}}$: one with a provable exponential lower bound on $\gamma_{\text{bw}}$ (Theorem~\ref{theorem:LB}), and another derived from practical applications (Example~\ref{example:health}). These cases demonstrate that high-threshold instances occur both theoretically and in practice, where standard methods are guaranteed to fail. In contrast, our approach successfully computes the BO policy even in these challenging settings, underscoring its robustness and practical utility.

\begin{theorem}\label{theorem:LB}
There exists an MDP \(M\) with $n$ states such that the threshold discount factor satisfies \(\gamma_{\text{bw}}^M \geq 1 - O(2^{-n/3})\).
\end{theorem}
\begin{proof}
    See Appendix~\ref{appendix:LB} for a construction.
\end{proof}

\begin{example}[Healthcare] \label{example:health}
    We examine a simplified model of an MDP used in healthcare~\citep{Bennett-Hauser,Clement-Petrik-II}, originally developed to simulate clinical decision-making using real patient data. The objective is to minimise patient mortality while also limiting the invasiveness of the prescribed drug dosage---low, medium, or high. The model consists of \(n\) states: the first \(n - 1\) represent progressively worsening health conditions, and the final state is an absorbing mortality state. The action set \(\{ \text{low}, \text{medium}, \text{high} \}\) corresponds to dosage levels. As shown in Figure~\ref{fig:mdp-health}, higher dosages increase the likelihood of recovery (i.e., transitions to earlier, healthier states). Rewards penalise more aggressive treatments: in states \(1\) through \(n-1\), the rewards are 10, 8, and 6 for low, medium, and high actions, respectively.

    Figure~\ref{fig:mdp-health} illustrates the family of MDPs we consider, and the accompanying table lists six representative instances with increasing state counts and their corresponding threshold discount factors \(\gamma_{\text{bw}}\). Notably, once the state count exceeds 25, the values of \(\gamma_{\text{bw}}\) surpass \(1 - 10^{-17}\), highlighting the numerical extremity of the problem.
\end{example}

\subsection{Key Idea}
We take policy iteration as our motivating method. At each step, given a current policy \(\pi\), we consider whether changing the action at some state \(s\) to an alternate action \(a\) would improve the policy. The standard approach evaluates this by comparing the value \(V^\pi(s)\) under the current policy with the \(Q\)-value \(Q^\pi(s,a)\), of taking action \(a\) and following \(\pi\) thereafter. Specifically, we check the sign of the difference:
\(
Q^\pi(s, a) - V^\pi(s).
\)

Our main idea is that we can determine the sign of this difference---i.e., whether switching to action \(a\) is beneficial---without computing the exact values of \(V^\pi(s)\) or \(Q^\pi(s, a)\). Instead, we treat both quantities as rational functions in the discount factor \(\gamma\), and focus on the sign of their difference as a function of \(\gamma\).

Let \(P(\gamma) = Q_\gamma^\pi(s,a) - V_\gamma^\pi(s)\). This is a rational function, and our goal is to determine the sign of \(P(\gamma)\) at a specific value of \(\gamma\), typically very close to 1 (e.g., \(\gamma = 0.999\)).

The key observation is that if we identify the last root of \(P(\gamma)\) before 1---say at \(\gamma = 0.9\)---then the sign of \(P(\gamma)\) remains constant in \((0.9, 1)\), since there are no sign changes beyond that point. Thus, we can infer the sign of \(P(0.999)\) by simply computing the sign of \(P(1)\), bypassing exact evaluations of \(V^\pi(s)\) or \(Q^\pi(s,a)\).

This insight enables a symbolic approach to policy improvement, replacing numerical evaluation of value function with algebraic comparisons yielding efficient algorithms even as \(\gamma \to 1\).

\begin{figure}[b]
    \vspace{-0.1cm}
  \centering
  \begin{tikzpicture}[scale=0.65,
    >={Stealth[length=6pt,width=7pt]},
    every edge/.append style={thick},
    node style/.style={draw, circle, minimum size=20pt, inner sep=0pt, align=center},
    red edge/.style={draw=DarkRed!60},
    blue edge/.style={draw=DarkBlue!70}
    ]
    \node[node style, fill=lightgray!20] (A) at (0,0) {$s_1$};
    \node[node style, fill=lightgray!20] (B) at (3,0) {$s_2$};

    \path[->, red edge] (A) edge node[midway, above, DarkRed] {\scriptsize $1$ }(B);
    \path[->, dashed, blue edge, loop, out=120, in=60, min distance=7mm] (A) edge node[above, DarkBlue] {\scriptsize $0.1$} (A);
    \path[->, red edge, loop, out=120, in=60, min distance=7mm] (B) edge node[above, DarkRed] {\scriptsize $0$} (B);
  \end{tikzpicture}

  \vspace{0.4cm}

  {\footnotesize
  \renewcommand{\arraystretch}{1.2}
  \begin{tabular}{|c|cccccccc|}
    \hline
    $u$ & $3$ & $4$ & $5$ & $6$ & $7$ & $8$ & $...$ & $17$ \\
    \hline
    VI Time (s) & 0.1 & 1.1 & 13 & 153 & 1828  & -- & $...$ & -- \\
    LP Status   & \tick & \tick & \tick & \tick & \tick & \cross & $...$ & \cross \\
    PI Status   & \tick & \tick & \tick & \tick & \tick & \tick & $...$ & \cross \\
    \hline
  \end{tabular}
  }
\vspace{0.3cm}
  \caption{\footnotesize Figure shows a deterministic MDP with two states: two actions (solid and dashed) are available from state 1, and a single action from state 2. Edge labels indicate the corresponding rewards. The accompanying table reports the performance of VI, LP, and PI as the discount factor \(\gamma = 1 - 10^{-u}\) increases. VI becomes impractically slow for \(u \geq 8\), LP produces incorrect results, and PI fails due to matrix singularities for \(u \geq 17\).}
  \label{fig:mdp-fail}
\end{figure}

\begin{figure}[b!]
\vspace{-0.1cm}
  \centering
   \begin{tikzpicture}[scale=0.8,
    >={Stealth[length=6pt,width=7pt]},
    every edge/.append style={thick},
    node style/.style={draw, circle, minimum size=20pt, inner sep=0pt, align=center},
    red edge/.style={draw=DarkRed!60},
    blue edge/.style={draw=DarkBlue!70}
    ]
    
    \node[node style, fill=lightgray!20] (A) at (0,0) {$s_1$};
    \node[node style, fill=lightgray!20] (B) at (1.75,0) {$s_2$};
    \node[node style, fill=lightgray!20] (C) at (3.5,0) {$s_3$};
    \node at (5.25,0) {\scriptsize $\cdots$}; 
    \node[node style, fill=lightgray!20] (D) at (7,0) {$s_{n-1}$};
    \node[node style, fill=lightgray!20] (E) at (7,-1.5) {$m$};

     \path[->, ] (D) edge node[right] {\scriptsize $t_7$ }(E);

    \path[->, bend left=45] (A) edge node[above] {\scriptsize $t_2$ }(B);
    \path[->, bend left=45] (B) edge node[above] {\scriptsize $t_5$ }(C);
    
    \path[->,dashed, bend left=45] (C) edge
        node[above, pos=0, xshift=10pt, yshift=10pt] {\scriptsize $t_5$}
        node[above, pos=1, xshift=-10pt, yshift=10pt] {\scriptsize $t_5$} (D);

    \path[->, bend left=45] (B) edge node[below] {\scriptsize $t_3$ }(A); 
    \path[->, bend left=45] (C) edge node[below] {\scriptsize $t_6$ }(B);

    \path[->, dashed, bend left=45] (D) edge
        node[below, pos=0, xshift=-10pt, yshift=-10pt] {\scriptsize $t_6$}
        node[below, pos=1, xshift=10pt, yshift=-10pt] {\scriptsize $t_6$} (C);

    \path[->, loop, out=120, in=60, min distance=7mm] (D) edge node[above, ] {\scriptsize $t_4$} (D);
    \path[->, loop, out=120, in=60, min distance=7mm] (B) edge node[above,] {\scriptsize $t_4$} (B);
    \path[->, loop, out=120, in=60, min distance=7mm] (A) edge node[above,] {\scriptsize $t_1$} (A);
    \path[->, loop, out=120, in=60, min distance=7mm] (C) edge node[above,] {\scriptsize $t_4$} (C);
    \path[->, loop, out=375, in=315, min distance=7mm] (E) edge node[right,] {\scriptsize $1$} (E);

  \node at (3.5,-3) { 
  \footnotesize
    \renewcommand{\arraystretch}{1.25}
    \begin{tabular}{|c|c|c|c|c|c|c|}
      \hline
      $n$ & 15 & 20 & 25& 30 & 35 & 40 \\ \hline
      $u$ & $9.98$ & $13.47$ & $16.97$ &$20.4$ & $23.96$ &$27.4$ \\ \hline
    \end{tabular}
  };

  \end{tikzpicture}
\vspace{0.3cm}
\caption{\footnotesize Family of healthcare-inspired MDPs~\cite{Clement-Petrik-II} along with their corresponding Blackwell threshold discount factors. The model uses seven distinct transition probabilities \(\{t_i\}_{i=1}^7\), across all states specified separately for each action:  
Low: \(\{0.7, 0.3, 0.3, 0.4, 0.3, 0.3, 0.3\}\);  
Medium: \(\{0.8, 0.2, 0.4, 0.4, 0.2, 0.4, 0.2\}\);  
High: \(\{0.9, 0.1, 0.5, 0.4, 0.1, 0.5, 0.1\}\).  
For example, taking the medium action from state \(s_3\) results in a transition to \(s_2\), \(s_3\), and \(s_4\) with probabilities \(0.4\), \(0.4\), and \(0.2\), respectively.  
The table reports values of \(u\) such that the Blackwell threshold satisfies \(\gamma_{\text{bw}} = 1 - 10^{-u}\), shown for increasing numbers of states \(n\).}

\label{fig:mdp-health}
\end{figure}

\section{Literature Review} \label{section:Lit Review}
\citet{blackwell1962} first demonstrated the existence of a Blackwell-optimal policy non-constructively. Building on this, \citet{miller-veinott} developed a policy iteration algorithm for finding the Blackwell-optimal policy, using the Laurent series expansion of \(V^\pi_\gamma\) around \(\gamma = 1\). Such expansions connect the gain and bias to the expected total discounted reward and are a standard tool for analysing finite-state undiscounted models. However, the method is computationally expensive, with only an exponential upper bound on its runtime. In Appendix~\ref{appendix:equivalence}, we show that this policy improvement procedure is equivalent to our rational function-based approach---both yield the same set of BO policies. Our algorithm, however, incorporates an additional step based on the Random-Facet method, which enables us to derive a subexponential bound.

\citet{veinott1969} introduced a new family of optimality criteria, known as \(N\)-discount optimality. For \(N = -1, 0, 1, \dots\), a policy \(\pi^\star\) is considered \(N\)-discount-optimal~\citep{veinott1969} if:  
\(
\lim_{\gamma \to 1} \ (1-\gamma)^{-N} [V_\gamma^{\pi^\star}(s) - V_\gamma^\pi(s)] \geq 0, \ \forall \pi, \forall s \in S.
\)  
This criterion captures the sensitivity of a policy’s optimality with respect to the parameter \(N\). Specifically, for \(N = -1\), the condition corresponds to gain optimality, and for \(N = 0\), it represents bias optimality. As \(N\) increases, the optimality condition becomes increasingly stringent, with Blackwell optimality being the most restrictive case.~\citet{veinott1969} showed that Blackwell optimality is equivalent to \(|S|\)-discount optimality.

As the sensitivity of the optimality criterion increases, designing efficient algorithms becomes more challenging. For $N = -1$ (gain optimality), the problem is efficiently solvable via policy improvement methods~\citep{howard1960} or linear programming~\citep{manne1960, derman1962, denardo1970}. The case $N = 0$ (bias optimality) was tackled by \citet{veinott1966} using policy improvement and by \citet{denardo1970computing} through linear programming. 

The fastest known algorithm for finding an \(N\)-optimal policy was developed by \citet{sullivan-veinott}. Their method decomposes the problem into a linear sequence of subproblems. Each subproblem is either a Maximum Transient Value (MTV) problem, which optimises short-term rewards, or a Maximum Reward Rate (MRR) problem, which maximises long-term average rewards. The input to each subproblem is determined by the solution of the previous one. Although each subproblem admits a linear programming formulation of size $poly(n,k)$ and can be solved in weakly polynomial time, the overall method is highly intricate and has no known implementations. In contrast, our algorithms are simple and yield strong complexity bounds---independent of the bit-size of rewards.

Building on the work of \citet{jeroslow1972}, \citet{hordijk-kallenberg} developed a method for comparing rational functions near zero, which they applied to discounted MDPs. They constructed a simplex tableau in which the entries are expressed as rational functions of the parameter $\rho = \frac{1 - \gamma}{\gamma}$. Using this symbolic representation, they applied Sturm’s Theorem to identify a threshold $\rho_0$ such that the current tableau remains optimal for all $\rho \geq \rho_0$. This threshold determines the next set of basic variables, allowing the tableau to be updated accordingly. The process is then repeated, successively identifying intervals $[\rho_1, \rho_0]$, until $\rho_0 = 0$, at which point the algorithm terminates with the optimal policy for the entire range of discount factors. Moreover, by setting $\rho = 0$ directly and adjusting the pivot selection rule, the method can be adapted to compute the Blackwell-optimal policy.
In contrast, our approach is simpler and integrates directly into the policy improvement framework, enabling generalisation to a broad class of efficient algorithms with provable upper bounds. 

\citet{Smallwood-1966} first introduced the concept of a threshold discount factor beyond which a discount-optimal policy becomes Blackwell-optimal. \citet{Clement-Petrik} demonstrated the existence of such a threshold for finite MDPs and provided an upper bound on \(1/(1 - \gamma_{\text{bw}})\), which is exponential in the number of states. However, no lower bound on this parameter was known. In this work, we provide the first exponential lower bound, showing that such thresholds can, in fact, be exponentially close to 1.

In a learning context, \citet{Mahadevan-1996a} introduced the first tabular Q-learning algorithm designed to achieve bias-optimal policies by optimising over the family of \(n\)-discount optimality criteria. \citet{dewanto-gallagher} presented a policy-gradient method for learning bias-optimal policies in unichain MDPs. Additionally, \citet{schneckenreither2020} proposed a model-free tabular algorithm for computing bias-optimal policies in unichain MDPs. 

\citet{boone-gaujal} studied the problem of identifying BO policies in deterministic MDPs within a fixed-confidence PAC-RL framework. They proved that this is impossible in general, unless the MDP satisfies uniqueness conditions on both the optimal cycle and the bias-optimal policy. For this maximal identifiable class, they proposed a sample-efficient algorithm based on generalised Bellman coefficients and structured confidence sets, achieving near-optimal bounds on the number of reward queries required.

\section{Method of Rational Functions} \label{section: Rational Fn}
In this section, we present the mathematical basis of our method for computing BO policies. We begin by defining an ordering of rational functions in the vicinity of 1. Using this ordering, we derive BO policies for both Deterministic MDPs (DMDPs) and general MDPs.

\subsection{Ordering of rational functions}
Consider two rational functions, \( r_1(x) = \frac{p_1(x)}{q_1(x)}, \quad r_2(x) = \frac{p_2(x)}{q_2(x)} \), where \(p_1, p_2, q_1, q_2\) are polynomials with real coefficients and $q_1, q_2 \not\equiv 0$. Let $\tau(x) = r_1(x) - r_2(x)$, we define $\tau \doteq 0$ if and only if $p_1(x)q_2(x) \equiv p_2(x)q_1(x)$. Suppose: \[\tau(x) = \frac{\eta(x)}{\delta(x)}=\frac{(1-x)^{c_1}\cdot \bar{\eta}(x)}{(1-x)^{c_2}\cdot \bar{\delta}(x)},\] where $c_1, c_2 \in \mathbb{Z}_{\ge 0}$ denote the multiplicities of the root $x=1$ in $\eta(x)$ and $\delta(x)$, respectively, and $\bar{\eta}(1)\neq 0, \bar{\delta}(1)\neq 0$. We define a total order \(\succ\) on the set of rational functions, such that:
\[ r_1 \succ r_2 \iff \bar{\eta}(1) \cdot \bar{\delta}(1)  > 0 .\]
We refer to this ordering as \(\mu\)-ordering. 

\begin{figure}
    \centering
    \begin{tikzpicture}[scale=0.65]
    \begin{axis}[
        axis lines = middle,
        xlabel = $x$,
        ylabel = {$r(x)$},
        xmin = 0.5, xmax = 1.25,
        ymin = -0.1, ymax = 0.5,
        grid = both,
        legend pos = north east,
        width = 0.5\textwidth,
        height = 0.3\textwidth
    ]
    
    \addplot [
        domain=0.6:1.1,
        samples=100,
        dashed,
        thick,
        DarkBlue,
    ] {(1-x)^2*(5*x-10)/(x-2)};
    \addlegendentry{$r_1(x)$}
    \addplot [
        domain=0.55:1.1,
        samples=100,
        solid,
        thick,
        DarkGreen,
    ] {(1-x)*(x-5)/(x-4)};
    \addlegendentry{$r_2(x)$}
    
    \draw[red, thick, dotted] (axis cs:1,-2) -- (axis cs:1,2);
    
    \end{axis}
    \end{tikzpicture}
    \vspace{0.3cm}
    \caption{\footnotesize Plot of the functions \(r_1\) and \(r_2\). Note that \(r_1(1) = r_2(1) = 0\), while \(r_1(1 - \epsilon) < r_2(1 - \epsilon)\) for all \(0 < \epsilon < \frac{1}{4}\). }
    \label{fig:plot}
\end{figure}
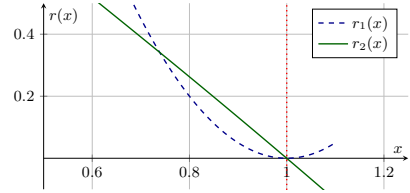

\begin{lemma}
    For finite rational functions $r_1$ and $r_2$, $r_1 \succ r_2$ $\iff$ $\exists \sigma \in (0, 1)$  such that $r_1(x) > r_2(x)$ $\forall x \in (\sigma,1)$.
\end{lemma}
\begin{proof}
    Since $\eta$ and $\delta$ are finite polynomials they have a finite number of roots. Let  $\sigma_1$ and $\sigma_2$ be the largest roots of $\eta$ and $\delta$ respectively in $(0,1)$ and let $\max\{\sigma_1,\sigma_2\}=\sigma$. It is clear that $\tau$ cannot change sign in $(\sigma,1)$. Since $\bar{\eta}(1)\neq 0$ and $\bar{\delta}(1) \neq 0$, the ratio $\bar{\tau}  = \frac{\bar{\eta}}{\bar{\delta}}$ must be finite and non-zero. As $\bar{\tau}$ is continuous we have: $\bar{\tau}(\sigma_0)>0 \iff \bar{\tau}(1)>0, \ \forall \sigma_0 \in (\sigma,1)$. Since $(1-\sigma_0)>0$, $\tau$ and $\bar{\tau}$ have the same sign at $\sigma_0$ for all $\sigma_0 \in (\sigma,1)$ and the result follows.    
\end{proof}
\begin{example}
    Let $r_1=\frac{(1-x)^2(5x-10)}{x-2}$ and $r_2=\frac{(1-x)(x-5)}{x-4}$. Then $\tau = \frac{(1-x)(-5x^2+24x-15)}{x-4}$ and $\bar{\eta}(1)\bar{\delta}(1)=4\cdot(-3)<0 \mathclose{\implies} r_2 \succ r_1$. (see Figure~\ref{fig:plot}).
\end{example}

\vspace{-0.25cm}

\subsection{Complexity}
The complexity of comparing two rational functions depends on two primary operations:  
\begin{itemize}
    \item Computing the difference \( \tau \): This involves subtracting one rational function from the other. Assume that the polynomials involved have a degree of \( O(d) \), the calculation of \( \tau \), which includes polynomial multiplication, can be performed in \( O(d \log d) \) steps using efficient polynomial multiplication algorithms. 
    \item Determining the multiplicity of 1 for \( \tau \): This requires evaluating the polynomial at 1 at most upto its $d$-th derivative. Using Horner's method, each evaluation at 1 takes \( O(d) \) operations resulting in $O(d^2)$ complexity to determine the multiplicity.
\end{itemize}
Therefore, the overall complexity of comparing two rational functions is \( O(d^2) \) operations.

Given a sequence having $t$ elements, finding a maximal/minimal element requires at most $t$ comparisons. Since each comparison between rational functions has a complexity of $O(d^2)$ under $\mu$-ordering, the overall complexity of computing the maximal or minimal rational function from the sequence is $O(td^2)$. We use $\widehat{\max},\widehat{\min}$ to denote such max and min operations respectively.

\subsection{Application} \label{subsection:application}
In the following sections, we consider four existing algorithmic families $\{ \mathcal{L}_i \}_{1 \leq i \leq 4}$, each generating a sequence of policies converging to an optimal policy $\pi^\star$ for a discounted MDP $M_\gamma$. The trajectory $\left( \pi^0_\gamma, \pi^1_\gamma, \dots, \pi^\star_\gamma \right)$ depends on both the MDP and $\gamma$. For each $\mathcal{L}_i$, we define a corresponding $\gamma$-independent algorithm $\mathcal{L}_i'$ that produces the same sequence for all sufficiently large discount factors. The following theorem establishes the existence of such a sequence.

\begin{theorem}
    There exists a $\tau \in [0,1)$ such that $\mathcal{L}_i$ follows the trajectory: $\left( \pi^0_\tau, \pi^1_\tau, \dots, \pi^\star_\tau \right)$ for all $\gamma \in [\tau,1)$.
\end{theorem}
\begin{proof}
    Let the discount factor $\gamma$ be treated as a symbolic variable. Then, equation~\eqref{value:cramer} defines a value vector $\mathbf{v}^\pi_\gamma$, whose components (one per state) are rational functions of $\gamma$.

    Define $f^{(i,j)}_s(\gamma) = V_\gamma^{\pi_j}(s) - V_\gamma^{\pi_i}(s)$, the value difference at state $s$ between policies $\pi_j$ and $\pi_i$. Since the MDP is finite, each $f^{(i,j)}_s(\gamma)$ is a ratio of finite-degree polynomials, and therefore has only finitely many roots. Let $\gamma^{(i,j)}_s$ denote the largest root of $f^{(i,j)}_s$ in the interval $[0,1)$. For any $\gamma > \gamma^{(i,j)}_s$, the sign of $f^{(i,j)}_s(\gamma)$ remains constant, meaning the relative ordering of $V_\gamma^{\pi_i}(s)$ and $V_\gamma^{\pi_j}(s)$ does not change beyond this point.
    Define the global threshold $$\tau = \max\limits_{s \in S,\, \pi_i,\pi_j \in \Pi} \gamma^{(i,j)}_s.$$ Then, for all $\gamma \in [\tau,1)$, the value ordering---and hence policy preferences---remains invariant across policy pairs in {$\Pi_\gamma$}.
\end{proof}

Since $\mu$-ordering provides an ordering or rational functions in the left-neighbourhood of 1, the theorem implies that $\mathcal{L}_i$ follows the same trajectory as when applied to $\mathcal{M}_{1-\epsilon}$  with $\epsilon \to 0$. Hence, both correctness and complexity guarantees of $\mathcal{L}_i$ in the discounted setting extend to the symbolic variant $\mathcal{L}'_i$, with only a polynomial overhead from symbolic evaluation.

\section{Efficient Planning with Blackwell Optimality} \label{section:Algos}
In this section, we describe algorithms of type $\mathcal{L}'_i$, identify the rational functions and corresponding thresholds that govern their policy trajectories, and analyse their computational complexity.
\subsection{Maximum Mean Weight Cycle} 
The algorithm of~\citet{madani2010discounted}, inspired by Karp’s method for finding the minimum mean‐weight cycle in a graph~\citep{karp1978}, is the fastest known approach for solving DMDPs under the discounted criterion.

The algorithm proceeds in two stages, both based on Bellman-Ford-style updates. In the first stage, it computes $d_i(s)$, the maximum discounted cost of an $i$-edge path starting from state $s$, for all states. Using these $d_i$ values, it evaluates
\[
y_0(s) = \min_{0 \leq i \leq n} \frac{d_n(s) - \gamma^{n-i}d_i(s)}{1 - \gamma^{n-i}},
\]  
for each $s$. In the second stage, the $y_0(s)$ values serve as initial state values for updating $y_i(s)$, and the maximum $y_i(s)$ over all states yields the optimal values.

The discounted costs of paths and the ratios \(y_i\) are rational functions of \(\gamma\). We apply $\mu$-ordering replacing the $\max\text{ and } \min$ operators with $\widehat{\max}\text{ and }\widehat{\min}$ respectively. Since the original algorithm focuses solely on computing values, we introduce additional variables \(\bar{\alpha}_i(s)\), \(\bar{y}_0(s)\), and \(\bar{a}_i(s)\) to track the actions taken during the updates of \(d_i(s)\), \(y_0(s)\), and \(y_i(s)\), respectively. This modification enables recovery of the BO policy. The full procedure is given in Algorithm~\ref{madani}.

Let $D^i_\gamma(s,a,s') = r(s,a,s')+\gamma d_{i-1}(s')$ and $Y^j_\gamma(s,a,s') = r(s,a,s')+\gamma y_{j-1}(s')$ where $d_i$ and $y_j$ are as defined in Algorithm~\ref{madani}. Then the threshold discount factor $\gamma_{1}$ is defined as:
\begin{multline*}
\gamma_1 \eqdef 
\sup_{\substack{1\le i\le n,\,1\le j\le n-1\\ s,s',s''\in S,\,a,a'\in A}} \
\inf_{\gamma\in[0,1)} 
\bigg\{\forall \tau\in(\gamma,1),\ \forall \Phi\in\{D^i,Y^j\}:\\
\Phi_\gamma(s,a,s')>\Phi_\gamma(s,a',s'') \Rightarrow
\Phi_\tau(s,a,s')>\Phi_\tau(s,a',s'')\bigg\}.
\end{multline*}

\begin{theorem}
    The DetMDP2-Blackwell procedure computes a BO policy for a DMDP with a runtime complexity of \(O(n^4k)\).
\end{theorem}
\begin{proof}
    As demonstrated earlier, the complexity of comparing two rational functions is \(O(n^2)\). Since the maximum is sought over the set of \(k\) actions, the complexity for finding the maximum at each step is \(O(n^2k)\). Given that the maximum is computed at most \(n^2\) times throughout the algorithm, the runtime complexity is \(O(n^4k)\).
\end{proof}
\begin{algorithm}
\caption{DetMDP2-Blackwell} \label{madani}
\scalebox{0.94}{
\begin{minipage}{\linewidth}
\begin{algorithmic}[1] 
\vspace{0.05cm}
\Procedure{$\Phi_D$}{$M=(S,A,T,r)$}
\For {each $s \in S$} 
    \State $d_0(s) \gets 0$
\EndFor
\For {$i=1$ to $n$}
    \For {each $s \in S$}
        \State $d_i(s)\gets \widehat{\max}_{s' \in S}  \ {r(s,a,s') + \gamma d_{i-1}(s')}$ 
        \State $\overline{\alpha}_i(s)\gets \argmaxhat_{a \in A} r(s,a,s') + \gamma d_{i-1}(s')$
    \EndFor
\EndFor

\For {each $s \in S$}
    \State $y_0(s)\gets \widehat{\min}_{0 \leq i \leq n} {\frac{d_n(s)- \gamma^{n-i}d_i(s)}{1-\gamma^{n-i}}}$ 
    \State $\overline{y}_0(s)\gets{\overline{\alpha}_n(s)}$
\EndFor

\For {$i=1$ to $n-1$}
    \For {each $s \in S$}
        \State $y_i(s)\gets \widehat{\max}_{s' \in S} \  {r(s,a,s') + \gamma y_{i-1}(s')}$ 
        \State $\overline{a}_i(s)\gets \argmaxhat_{a \in A}  r(s,a,s') + \gamma y_{i-1}(s')$
    \EndFor
\EndFor

\For {each $s \in S$}
    \State $\overline{i}_s\gets \argmaxhat_{0 \leq i \leq n} {y_i(s)}$ 
    \State $\pi^*_{\text{bw}}(s) \gets \overline{a}_{\overline{i}_s}(s)$
\EndFor
\State \textbf{return} $\pi^*_{\text{bw}}$
\EndProcedure
\vspace{0.05cm}
\end{algorithmic}
\end{minipage}
}
\end{algorithm}

\subsection{Max Gain} \label{subsection:MGS}
The Max-Gain Simplex (MGS) algorithm~\citep{post2013simplex} computes the optimal policy for DMDPs in strongly polynomial time. For a DMDP with a discount factor \(\gamma\), the algorithm iteratively selects the state-action pair with the highest gain to transition to a new policy. Specifically, starting from a policy \(\pi\), it transitions to a new policy \(\overline{\pi}\) defined as:  
\[
\overline{\pi}(s) = 
\begin{cases} 
\pi(s), & \text{if } s \neq \overline{s}, \\ 
\overline{a}, & \text{if } s = \overline{s}, 
\end{cases}
\]  
where \((\overline{s}, \overline{a})\) is chosen to satisfy:  
\[
(\overline{s}, \overline{a}) = \argmax_{(s, a)} \left(Q^\pi(s, a) - V^\pi(s)\right).
\]  

\subsubsection{Computing BO policies}
Our algorithm for computing BO policies mirrors the procedure of MGS, but with an updated max operator. From \eqref{bellman-V} and \eqref{bellman-Q}, it is clear that the quantity \(Q_\gamma^\pi(s, a) - V_\gamma^\pi(s)\) is a rational function of \(\gamma\). Using $\mu$-ordering, we redefine the max operator as:  
\[
(\overline{s}, \overline{a}) = \argmaxhat_{(s, a)} \left(Q_\gamma^\pi(s, a) - V_\gamma^\pi(s)\right).
\]  
\begin{theorem}
    The described procedure computes a BO policy for a DMDP in at most \(O(n^7k^3\log^2n)\) iterations.
\end{theorem}
\begin{proof}
    The MGS algorithm completes in at most \(O(n^5 k^2 \log^2 n)\) iterations~\citep{post2013simplex}. Our algorithm introduces an additional \(O(n^2k)\) steps per iteration to determine the max-gain. Hence, the total computational complexity for obtaining a BO policy is \(O(n^7 k^3 \log^2 n)\).
\end{proof}
Since MGS and the subsequent algorithms compute Q-values to guide action selection, their threshold discount factor is $\gamma_Q$.

\subsection{Random-Facet} \label{subsection:RF}
~\citet{MSW} introduced the randomised pivot rule Random-Facet, which gives an upper bound of $2^{\sqrt{n \log m}}$ on the expected number of pivot steps to solve any linear program with $n$ variables and $m$ constraints. Combining the algorithm with Clarkson's method~\citep{Clarkson95} yields the tightest known bound of 
$O\left(n^2m + e^{O(\sqrt{n \log n})}\right)$.

Let $p = (s, a)$ be a state-action pair, and define $f^\pi(s, a) = Q^\pi(s, a) - V^\pi(s)$. The pair $p$ belongs to $\pi$ if $\pi(s) = a$, and is \textit{improving} for $\pi$ iff $f^\pi(s, a) > 0$. The Random-Facet algorithm relies on determining whether a given pair $p$ is improving for a policy $\pi$, which amounts to checking the sign of $f^\pi(s, a)$ at each step.

The algorithm can be adapted to MDPs as follows: given an MDP $M$ and an initial policy $\pi$, select a state-action pair $p \notin \pi$, and recursively solve $M$ without $p$, yielding a new policy $\pi'$. If $p$ is not improving for $\pi'$, then $\pi'$ is guaranteed to be optimal. Otherwise, update $\pi'$ by switching $p$, and repeat the process.

Let $\hat{f}^\pi_\gamma(s,a) = Q^\pi_\gamma(s,a) - V^\pi_\gamma(s)$ denote the rational function in $\gamma$. Our algorithm for identifying a BO policy adapts the Random-Facet procedure, applying $\mu$-ordering on $\hat{f}$ to determine its sign in the vicinity of $1$. The pseudocode is given in Algorithm~\ref{RF}. We denote the set of state-action pairs corresponding to $M$ and $\pi$ by $M_p$ and $\pi_p$, respectively.

\begin{algorithm} 
\caption{Random-Facet-Blackwell} \label{RF}
\scalebox{0.94}{ 
\begin{minipage}{\linewidth}
\begin{algorithmic}[1] 
\vspace{0.05cm}
\Procedure{$\Phi_R$}{$M_p,\pi$}
\If {$\hat{f}^{\pi}(p) \preceq 0, \forall p\in M_p$}
    \State $\pi^\star_{\text{bw}} \gets \pi$
    \State \textbf{return} $\pi^\star_{\text{bw}}$
\Else
    \State Pick $p \in M_p \setminus \pi_p$ uniformly at random
    \State $\pi' \gets \Phi_R(M_p \setminus p,\pi)$ \Comment{\textcolor{gray}{1st call}}
    \If {$\hat{f}^{\pi'}(p)\succ 0$}
        \State $\pi^{\prime\prime} \gets switch(p,\pi')$
        \State \textbf{return} $\Phi_R(M_p,\pi^{\prime\prime})$ \Comment{\textcolor{gray}{2nd call}}
    \Else
        \State \textbf{return} $\pi'$
    \EndIf
\EndIf
\EndProcedure
\vspace{0.05cm}
\end{algorithmic}
\end{minipage}

}
\end{algorithm}

\begin{theorem}
  The Random-Facet-Blackwell algorithm computes a BO policy for an MDP in at most \(poly(n, k) \cdot e^{O(\sqrt{n \log n})}\) expected iterations.
\end{theorem}
\begin{proof}
    The Random-Facet algorithm, when combined with Clarkson's algorithm, has an expected runtime of \(O(n^3k + e^{O(\sqrt{n \log n})})\)~\citep{MSW}. Our algorithm introduces an additional \(n^2\) operations per recursive call to determine the sign of \(\hat{f}^\pi\). Consequently, the overall runtime complexity increases by a polynomial factor in \(n\), resulting in the desired bound.    
\end{proof}

\subsection{Generalisation: PI improvement}
The approaches in Sections~\ref{subsection:MGS} and \ref{subsection:RF} extend naturally to any Policy Improvement (PI) procedure.
A generic PI procedure operates as follows: given a policy $\pi$, define the set of improving state-action pairs: 
$
J^{\pi} = \{ (s,a) \mid Q^\pi(s,a) > V^\pi(s) \}.
$
At each iteration, select a subset $\Theta \subseteq J^{\pi}$ containing at most one action per state, and construct a new policy $\pi'$ such that $\pi'(s) = a$ for all $(s,a) \in \Theta$. The rule for selecting $\Theta$ is algorithm-specific and crucial to the procedure’s complexity. For example, in the Max-Gain method, $\Theta$ consists of the single pair $(s,a)$ that maximises: $Q^\pi(s,a) - V^\pi(s)$.

As before, $Q_\gamma^\pi(s,a) - V_\gamma^\pi(s)$ is a rational function in $\gamma$, whose sign can be determined via $\mu$-ordering. This leads to a PI procedure for computing BO policies in MDPs. We examine the three tightest known variants of \textit{memoryless} PI algorithms---$\mathcal{A}_1$, $\mathcal{A}_2$, and $\mathcal{A}_3$---with respective bounds $\mathcal{B}_1$, $\mathcal{B}_2$, and $\mathcal{B}_3$, each depending only on $n$ and $k$ under the discounted criterion. Applying $\mu$-ordering, we generalise each variant to achieve bounds of $\mathrm{poly}(n,k)\cdot \mathcal{B}_i$ under the Blackwell criterion. Below, we summarise $\mathcal{A}_1$, $\mathcal{A}_2$, and $\mathcal{A}_3$ together with their corresponding bounds.

\begin{itemize}
    \item $\mathcal{A}_1$: Batch‐Switching policy iteration~\cite{BSPI} partitions the state space into batches with a fixed ordering and, at each step, switches states from $J^\pi$ in decreasing order of their batch index, yielding an iteration bound of $\mathcal{B}_1 = O(1.64^n)$. 

    \smallskip
    
    \item $\mathcal{A}_2$: Howard's policy iteration~\cite{howard1960} is a greedy procedure that, at each iteration, switches every improvable state---i.e., it maximises the set \(\Theta\). While highly efficient in practice, its theoretical upper bound remains exponential in \(n\), specifically $\mathcal{B}_2= O(\frac{k^n}{n})$~\cite{mansour-singh}.

    \smallskip
    
    \item $\mathcal{A}_3$: Randomised Simple policy iteration~\cite{RSPI} assumes an indexing of the states. At each iteration, it considers all improving pairs whose state has the highest index, and switches exactly one, chosen uniformly at random. Its iteration bound is $\mathcal{B}_3 = O((\log k)^n)$.

\end{itemize}

\section{Summary and Discussion} \label{section:Summary}
In this paper, we used an ordering of rational functions near $1$ to develop novel and efficient algorithms for computing BO policies in both MDPs and DMDPs. Our methods attain the tightest known complexity bounds, advancing the state of the art through a simple, theoretically robust framework for BO policy computation.

To illustrate the limitations of existing dynamic programming algorithms, we presented two examples, one of which gives an exponential lower bound on the threshold $\gamma_{\text{bw}}$. This bound not only underscores the complexity of computing BO policies but also limits the generalisability of existing proof techniques. In particular, \citet{mukherjee-kalyanakrishnan} prove a subexponential upper bound for Howard’s policy iteration in DMDPs by analysing ratios of polynomials derived from Q-value comparisons and bounding the roots near 1 to control \(\gamma_Q\). Our exponential lower bound on \(\gamma_{\text{bw}} \leq \gamma_Q\) shows that such techniques cannot extend to the stochastic case.

We also implemented and tested a basic, unoptimised version of our symbolic policy iteration~\cite{blackwell-pi-code}. For DMDPs, the method is reasonably efficient, solving instances with up to 100 states in about three minutes on a standard desktop (AMD Ryzen 7 5700G, 16 GB RAM). For general MDPs, runtime increases more sharply with stochasticity due to the cost of symbolic matrix inversion: problems with up to 10-15 states solve within minutes, whereas 20-state instances may require over an hour. Benchmarking against prior BO algorithms is challenging---few exist, and those that do are rarely presented in a form amenable to straightforward implementation. This scarcity of practical baselines underscores the need for simple, implementable approaches such as ours. Although our current results serve primarily as a proof of concept, we plan to optimise and scale up the implementation in future work.

While our focus has been on planning with full model knowledge, the structural insights and algebraic techniques developed here may also inform learning-based approaches. In particular, model-based reinforcement learning could leverage efficient BO policy computation once an approximate model is inferred from data. However, as recent work has shown~\citep{boone-gaujal}, identifying BO policies with limited samples is generally impossible without strong assumptions. A promising direction for future research is to investigate whether our algebraic characterisations can help delineate the class of MDPs where reliable identification is feasible, or guide the design of robust algorithms under uncertainty.


\clearpage
\bibliography{references}
\clearpage
\appendix
\onecolumn

\section{$\gamma_{\text{bw}}$ : A Lower Bound}\label{appendix:LB}
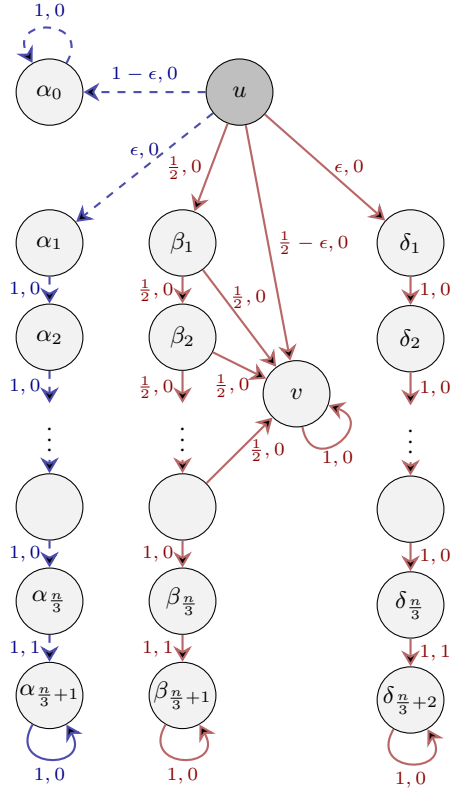
\begin{figure}[b]
  \centering
    \begin{tikzpicture}[scale=1,
    >={Stealth[length=6pt,width=7pt]},
    every edge/.append style={thick},
    node style/.style={draw, circle, minimum size=25pt, inner sep=0pt, align=center},
    red edge/.style={draw=DarkRed!60},
    blue edge/.style={draw=DarkBlue!70}
    ]
    
    \node[node style, fill=lightgray] (A0) at (2.5,2) {$u$};

    \node[node style, fill=lightgray!20] (B0) at (0,2) {$\alpha_0$};
    \path[->, blue edge, dashed, loop, out=60, in=120, min distance=7mm] (B0) edge node[above, DarkBlue] {\scriptsize $1,0$} (B0);
    
    \node[node style, fill=lightgray!20] (A1) at (0,0) {$\alpha_1$};
    \node[node style, fill=lightgray!20] (B1) at (0,-1.25) {$\alpha_2$};
    \node[draw=none] (dots1) at (0,-2.5) {$\vdots$};
    \node[node style, fill=lightgray!20] (C1) at (0,-3.5) {};
    \node[node style, fill=lightgray!20] (D1) at (0,-4.75) {$\alpha_{\frac{n}{3}}$};
    \node[node style, fill=lightgray!20] (E1) at (0,-6) {$\alpha_{\frac{n}{3}\mathclose{+}1}$};

    \path[->, blue edge, dashed] (A1) edge node[midway, left, DarkBlue] {\scriptsize $1, 0$ }(B1);
    \path[->, blue edge, dashed] (B1) edge node[midway, left, DarkBlue] {\scriptsize $1, 0$ }(dots1);
    \path[->, blue edge, dashed] (dots1) edge (C1);
    \path[->, blue edge, dashed] (C1) edge node[midway, left, DarkBlue] {\scriptsize $1, 0$ }(D1);
    \path[->, blue edge, dashed] (D1) edge node[midway, left, DarkBlue] {\scriptsize $1, 1$ }(E1);

    \node[node style, fill=lightgray!20] (A2) at (1.75,0) {$\beta_1$};
    \node[node style, fill=lightgray!20] (B2) at (1.75,-1.25) {$\beta_2$};
    \node[draw=none] (dots2) at (1.75,-2.5) {$\vdots$};
    \node[node style, fill=lightgray!20] (C2) at (1.75,-3.5) {};
    \node[node style, fill=lightgray!20] (D2) at (1.75,-4.75) {$\beta_{\frac{n}{3}}$};
    \node[node style, fill=lightgray!20] (E2) at (1.75,-6) {$\beta_{\frac{n}{3}\mathclose{+}1}$};

    \path[->, red edge] (A2) edge node[midway, left, DarkRed!100] {\scriptsize $\frac{1}{2}, 0$ }(B2);
    \path[->, red edge] (B2) edge node[midway, left, DarkRed] {\scriptsize $\frac{1}{2}, 0$ }(dots2);
    \path[->, red edge] (dots2) edge (C2);
    \path[->, red edge] (C2) edge node[midway, left, DarkRed] {\scriptsize $1, 0$ }(D2);
    \path[->, red edge] (D2) edge node[midway, left, DarkRed] {\scriptsize $1,1$ }(E2);

    \node[node style, fill=lightgray!20] (A3) at (3.25,-2) {$v$};
    \path[->, red edge, loop, out=280, in=340, min distance=7mm] (A3) edge node[below, DarkRed] {\scriptsize $1,0$} (A3);
    
    \path[->, red edge] (A2) edge node[midway, xshift=-7pt, yshift=8pt, right, DarkRed] {\scriptsize $\frac{1}{2}, 0$ }(A3);
    \path[->, red edge] (B2) edge node[midway, xshift=-3pt, below, DarkRed] {\scriptsize $\frac{1}{2}, 0$ } (A3);
    \path[->, red edge] (C2) edge node[midway, right, DarkRed] {\scriptsize $\frac{1}{2}, 0$ }(A3);

    \node[node style, fill=lightgray!20] (A4) at (4.75,0) {$\delta_1$};
    \node[node style, fill=lightgray!20] (B4) at (4.75,-1.27) {$\delta_2$};
    \node[draw=none] (dots4) at (4.75,-2.53) {$\vdots$};
    \node[node style, fill=lightgray!20] (C4) at (4.75,-3.53) {};
    \node[node style, fill=lightgray!20] (D4) at (4.75,-4.8) {$\delta_{\frac{n}{3}}$};
    \node[node style, fill=lightgray!20] (E4) at (4.75,-6.06) {$\delta_{\frac{n}{3}\mathclose{+}2}$};

    \path[->, red edge] (A4) edge node[midway, right, DarkRed] {\scriptsize $1, 0$ }(B4);
    \path[->, red edge] (B4) edge node[midway, right, DarkRed] {\scriptsize $1, 0$ } (dots4);
    \path[->, red edge] (dots4) edge (C4);
    \path[->, red edge] (C4) edge node[midway, right, DarkRed] {\scriptsize $1, 0$ }(D4);
    \path[->, red edge] (D4) edge node[midway, right, DarkRed] {\scriptsize $1, 1$ }(E4);

    \path[->, blue edge, dashed] (A0) edge node[midway, above, DarkBlue] {\scriptsize $1-\epsilon, 0$ }(B0);
    \path[->, blue edge, dashed] (A0) edge node[midway, above, DarkBlue] {\scriptsize $\epsilon, 0$ }(A1);
    \path[->, red edge] (A0) edge node[midway, left, DarkRed] {\scriptsize $\frac{1}{2}, 0$ }(A2);
    \path[->, red edge] (A0) edge node[midway, right, DarkRed] {\scriptsize $\frac{1}{2}-\epsilon, 0$ }(A3);
    \path[->, red edge] (A0) edge node[midway, right, DarkRed] {\scriptsize $\epsilon, 0$ }(A4);

    \path[->, blue edge, loop, out=240, in=300, min distance=7mm] (E1) edge node[below, DarkBlue] {\scriptsize $1,0$} (E1);

    \path[->, red edge, loop, out=240, in=300, min distance=7mm] (E2) edge node[below, DarkRed] {\scriptsize $1,0$} (E2);

    \path[->, red edge, loop, out=240, in=300, min distance=7mm] (E4) edge node[below, DarkRed] {\scriptsize $1,0$} (E4);

    \end{tikzpicture}
\caption{MDP with states {$u,v,\alpha_i,\beta_i,\delta_i$}. State $u$ has 2 available actions $a_0$ (dashed) and $a_1$ (solid) and the other states have a single action. The edges are labelled: $T(s,a,s'), R(s,a,s')$ and $0<\epsilon<\frac{1}{2}$.}
\label{fig:lb} 
\end{figure}

\vspace{0.5cm}
In this section, we construct an MDP with a threshold discount factor $\gamma_{\text{bw}}$ that is exponentially close to 1, as established in Theorem~\ref{theorem:LB}.
\subsection{Construction}
Consider the MDP shown in Figure~\ref{fig:lb}, consisting of $n+7$ states. From the initial state \(s_0 = u\), there are two available actions: \(a_0\) and \(a_1\). All other states allow only a single action \(a_0\). There are two policies: $\pi_0 = a_0a_0^{n-1}$ and $\pi_1=a_1a_0^{n-1}$. The value of \(\pi_0\) from the start state is:
\begin{align*}
V^{\pi_0} \left( u \right) &= \gamma \left[ (1-\epsilon) V^{\pi_0} \left( \alpha_0 \right) + \epsilon V^{\pi_0}\left( \alpha_1 \right) \right]  \\
& = \epsilon \gamma^ {\frac{n}{3}} \left[ 1 +  V^{\pi_0} \left( \alpha_{\frac{n}{3}\mathclose{+} 1} \right) \right] \\ 
& =  \epsilon \gamma^{\frac{n}{3}} 
\end{align*}
And the Q-value of taking action \(a_1\) from the same state under policy \(\pi_0\) is given by:
\begin{align*}
Q^{\pi_0} \left( u ,a_1 \right) &= \gamma \left[ \frac{1}{2} V^{\pi_0} \left(\beta_1 \right) + \left(\frac{1}{2} - \epsilon \right) V^{\pi_0}\left( v \right)  + \epsilon V^{\pi_0}\left( \delta_1 \right) \right]  \\
& = \gamma^ {\frac{n}{3}} \left(\frac{1}{2} \right) ^{\frac{n}{3}}\left[ 1 +  V^{\pi_0} \left(\beta_ {\frac{n}{3}\mathclose{+} 1} \right) \right]  \\ 
& \quad \quad \quad \quad \quad \quad +   \epsilon \gamma^ {\frac{n}{3}+1} \left[ 1 +  V^{\pi_0} \left(\delta_{ \frac{n}{3}\mathclose{+} 2} \right) \right]\\ 
& =  \gamma^{\frac{n}{3}} \left(\frac{1}{2}\right)^{\frac{n}{3}} + \epsilon \gamma^{\frac{n}{3}+1} 
\end{align*}
Therefore:
\[ Q^{\pi_0} \left( u,a_1 \right) > V^{\pi_0}\left( u\right) \implies \epsilon \gamma^{\frac{n}{3}} \left( \frac{1}{\epsilon 2^{\frac{n}{3}}} + \gamma -1 \right) > 0\]
This implies that action \(a_1\) is better than \(a_0\) in state $u$ whenever
\[
\gamma > \gamma_0 = 1 - \tfrac{1}{\epsilon 2^{n/3}}.
\]
 Therefore $\pi_1$ remains Blackwell-optimal beyond the threshold $\gamma_0$. This gives a lower bound $\gamma_{\text{bw}} \geq 1 - O(2^{-n/3})$.

\section{Equivalence of PI procedures} 
\label{appendix:equivalence}
We present the policy improvement procedure of~\citet{miller-veinott} and show that its sequence of visited policies matches that of our rational function-based approach. Our method further integrates the Random-Facet algorithm to obtain a subexponential bound. The steps for computing the difference vector $\mathbf{q}^\pi_a - \mathbf{v}^\pi$ for a given policy $\pi$ and action $a$ are outlined in Algorithm~\ref{algorithm:veinott}. 

\begin{algorithm}[H]
\caption{Policy Improvement}\label{algorithm:veinott}
\begin{algorithmic}[1]
\Procedure{Compute\_Q\_minus\_V}{$a, \pi, \mathcal{M}$}
    \State Compute $P^\pi$ and $\mathbf{r}^\pi$
    \State $P^* \gets \underset{\rho \to 0}{\lim} \, \rho \left(\rho I - (P^\pi - I)\right)^{-1}$
    \State $D^\pi \gets (I - P^*)(P^* - (P^\pi - I))^{-1}$
    
    \State Define $\mathbf{y}_j^\pi$ as:
    \[
    \mathbf{y}_j^\pi =
    \begin{cases}
        P^* \mathbf{r}^\pi, & \text{if } j = -1, \\
        (-1)^j \left(D^\pi\right)^{j+1} \mathbf{r}^\pi, & \text{if } j \geq 0
    \end{cases}
    \]
    
    \State Define $\mathbf{\psi}_j^{a, \pi}$ as:
    \[
    \psi_j^{a, \pi} =
    \begin{cases}
        \left(P_a - I\right) \mathbf{y}_j^\pi, & \text{if } j = -1, \\
        \mathbf{r}_a + \left(P_a - I\right) \mathbf{y}_j^\pi - \mathbf{y}_{j-1}^\pi, & \text{if } j = 0, \\
        \left(P_a - I\right) \mathbf{y}_j^\pi - \mathbf{y}_{j-1}^\pi, & \text{if } j \geq 0
    \end{cases}
    \]

    \State Construct the matrix $\Psi$:
    \[
    \Psi = \begin{bmatrix}
        \psi_{-1}^{a, \pi} & \psi_0^{a, \pi} & \cdots & \psi_{|S|}^{a, \pi}
    \end{bmatrix}
    \]

    \State Identify the first non-zero column of $\Psi$.
    \State Positive rows (states) in this column correspond to states $s$ such that $Q^\pi(a,s) > V^\pi(s)$.
\EndProcedure
\end{algorithmic}
\end{algorithm}

\begin{theorem}
    Let \(\mathbf{\Delta}^\pi(\gamma) = \mathbf{q}^\pi_a(\gamma) - \mathbf{v}^\pi(\gamma)\).
    For a fixed state \(s\in S\), write
    \(
    \mathbf{\Delta}^\pi_s(\gamma) = \frac{A(\gamma)}{B(\gamma)},
    \)
    with polynomials \(A,B\). Factor out the multiplicities of the root \(\gamma=1\):
    \(
    A(\gamma)=(1-\gamma)^{c_1}A_1(\gamma),\
    B(\gamma)=(1-\gamma)^{c_2}B_1(\gamma),
    \)
    where \(c_1,c_2\in\mathbb{Z}_{\ge 0}\) and \(A_1(1)\neq 0,\ B_1(1)\neq 0\).
    Define \(z_s(\gamma)=\frac{A_1(\gamma)}{B_1(\gamma)}\)
    Then,
    \(
    z_s(1)=\psi^{a,\pi}_{j_0}(s), \ j_0=\min\{j\mid \psi^{a,\pi}_j(s)\neq 0\}
    \)
    where $\psi^{a,\pi}$ is defined in Algorithm~\ref{algorithm:veinott}.
\end{theorem}

\begin{proof}
    The Laurent series expansion of the value function of a policy $\pi$ and discount factor $\gamma$ is given by~\citep{puterman}:
    \begin{equation} \label{laurent-series}
        \mathbf{v}^\pi_\gamma = (1+\rho)\sum_{j=-1}^\infty{\rho^j \mathbf{y}_j}
    \end{equation}
    where $\rho=\frac{1-\gamma}{\gamma}$, and $\mathbf{y}_j=\mathbf{y}^\pi_j$ as defined in Algorithm~\ref{algorithm:veinott}. \\
    Now consider the term: 
    \[\mathbf{q}^\pi_a- \mathbf{v}^\pi = \mathbf{r}_a+ \left( \gamma P_a - I\right)\mathbf{v}^\pi\]
    Replacing $\mathbf{v}^\pi$ from \eqref{laurent-series} we get:
    \begin{align*}
        \mathbf{q}^\pi_a- \mathbf{v}^\pi & =  \mathbf{r}_a + \left[P_a- (1+\rho) I \right] \sum_{j=-1}^\infty{\rho^j \mathbf{y}_j} \\
        & = \mathbf{r}_a + \sum_{j=-1}^\infty{\rho^j \left[P_a - I - \rho I \right] \mathbf{y}_j} \\
        & = \mathbf{r}_a  + \sum_{j=-1}^\infty{\rho^j \left(P_a - I \right) \mathbf{y}_j   -  \rho^{j+1} \mathbf{y}_j} \\
        & = \rho^{-1}(P_a-I)\mathbf{y}_{-1} + \left[\mathbf{r}_a+(P_a - I) \mathbf{y}_0 - \mathbf{y}_{-1}\right] 
        + \sum_{j=1}^\infty{\rho^j \left[(P_a - I)\mathbf{y}_j - \mathbf{y}_{j-1}\right]} \\
        & = \sum_{j=-1}^\infty{\rho^j \psi^{a,\pi}_j }
    \end{align*}
    Again consider the term \(\mathbf{q}^\pi_a- \mathbf{v}^\pi\) using equation~(\ref{poly-Q}). We have:
    \[\mathbf{q}^\pi_a- \mathbf{v}^\pi= \mathbf{r}_a + \left( \gamma P_a - I \right) \frac{\mathbf{n}_\pi}{{d}_\pi}\]
    Thus each entry of \(\mathbf{q}^\pi_a- \mathbf{v}^\pi\) is a ratio of two polynomials say $A$ and $B$. Write $A(\gamma) = (1-\gamma)^{m_1}A_1(\gamma)$ and $B(\gamma) = (1-\gamma)^{m_2} B_1(\gamma)$ where $m_1,m_2$ are integers greater than or equal to zero such that $A_1(1)\neq 0 \ \land \ B_1(1) \neq 0$. Now consider the function:
    \[f(\gamma)=\frac{A(\gamma)}{B(\gamma)} = (1-\gamma)^{m_1-m_2} \frac{A_1(\gamma)}{B_1(\gamma)}.\] 
    Let $t=m_1-m_2$ and $z(\gamma)=\frac{A_1(\gamma)}{B_1(\gamma)}$, then we have two cases: \\
    Case 1: $t \geq 0$ \\
    The Laurent series of $f$ around $\gamma=1$ is of the form:
    \[c_{t} (1-\gamma)^t + c_{t+1}(1-\gamma)^{t+1} +...\]
    It is clear that \(c_t = \frac{f^{(t)}(1)}{t!} = z(1)\). \\
    Case 2: $t < 0$ \\ 
    The Laurent series of $f$ around $\gamma=1$ is of the form:
    \[\frac{c_t}{(1-\gamma)^{-t}} + \frac{c_{t+1}}{(1-\gamma)^{-(t+1)}}  +...\]
    Here we have: $c_{t} = \lim_{\gamma \to 1} (1-\gamma)^{-t} f(\gamma) = z(1)$. \\
    Therefore the first non-zero term of the Laurent series expansion of \(\mathbf{q}^\pi_a- \mathbf{v}^\pi\) at state $s$ is given by: $\psi_{j_0}^{a,\pi} (s)= z(1)$, where $j_0$ is the index of the first non-zero term of $\psi^{a,\pi}$.
\end{proof}


\end{document}